\documentclass[runningheads]{llncs}
\usepackage{graphicx}
\usepackage{booktabs}
\usepackage{algorithm}
\usepackage{algorithmic}
\usepackage{multirow}
\usepackage{amsfonts}
\usepackage{url}
\usepackage{amsmath,amssymb}
\DeclareMathOperator{\len}{len}
\DeclareMathOperator{\D}{\mathcal{D}}
\DeclareMathOperator*{\argmax}{arg\,max}
\DeclareMathOperator{\arc}{Arc}
\DeclareMathOperator{\MST}{MST}
\DeclareMathOperator{\FF}{FF}
\DeclareMathOperator{\BN}{BN}
\DeclareMathOperator{\MHA}{MHA}
\DeclareMathOperator{\opt}{opt}
\newcommand*{\tran}{^{\mkern-1.5mu\mathsf{T}}}
\renewcommand{\vec}[1]{\mathbf{#1}}
\begin{document}

\title{Deep-Steiner: Learning to Solve the Euclidean Steiner Tree Problem}
\author{Siqi Wang\orcidID{0000-0002-0089-1005} \and
Yifan Wang\orcidID{0000-0002-6312-4663} \and
Guangmo Tong\orcidID{0000-0003-3247-4019}}
%\author{Submitted for blind review}
%
\authorrunning{Wang et al.}
\institute{University of Delaware, Newark DE 19716, USA \\
\email{\{wsqbit,yifanw,amotong\}@udel.edu}}
%\institute{}
%
\maketitle              % typeset the header of the contribution
\begin{abstract}
The Euclidean Steiner tree problem seeks the min-cost network to connect a collection of target locations, and it underlies many applications of wireless networks. In this paper, we present a study on solving the Euclidean Steiner tree problem using reinforcement learning enhanced by graph representation learning. Different from the commonly studied connectivity problems like travelling salesman problem or vehicle routing problem where the search space is finite, the Euclidean Steiner tree problem requires to search over the entire Euclidean space, thereby making the existing methods not applicable. In this paper, we design discretization methods by leveraging the unique characteristics of the Steiner tree, and propose new training schemes for handling the dynamic Steiner points emerging during the incremental construction. Our design is examined through a sanity check using experiments on a collection of datasets, with encouraging results demonstrating the utility of our method as an alternative to classic combinatorial methods.

\end{abstract}
\section{Introduction}
Network connection is an important issue in wireless networks \cite{fujiwara2004hybrid,caro2005wireless,obayiuwana2017network}, and a key problem in such studies is to build a connected network with the minimum cost, for example,  the location-selection problem \cite{lu2011location}, the relay node placement problem \cite{cheng2008relay}, and the network connectivity restoring problem \cite{senel2011relay}. Such problems can often be nicely reduced to the classic Euclidean Steiner tree (EST) problem, where we seek to connect a collection of points in the Euclidean space using Steiner minimal trees. The EST problem is well-known to be NP-hard, even in the two-dimensional Euclidean space \cite{garey1977complexity}, and various algorithms have been developed for effective and efficient solutions \cite{hwang1992steiner,winter1997euclidean,juhl2018geosteiner,Thompson1973TheMO,smith1981n,arora1998polynomial,dreyer1998two}.
Recently, a trending direction is to design reinforcement learning diagrams for solving combinatorial optimization problems in a data-driven manner. This is driven by as least two compelling reasons: philosophically, we are wondering if effective heuristics can be automatically inferred by artificial intelligence \cite{nazari2018reinforcement}; practically, heuristics produced by reinforcement learning are tuned by the empirical distribution of the inputs, and can therefore offer better performance for a specific application scenario \cite{Bello2017NeuralCO} \cite{nazari2018reinforcement}. In this paper, we revisit the EST problem and design reinforcement learning methods.

\textbf{Challenges.} Learning-based methods have recently shown promising results in solving combinatorial optimization problems \cite{Bello2017NeuralCO,khalil2017learning,kool2018attention}. In principle, searching the best solution to combinatorial optimization problems can be modeled as a Markov decision process (MDP) in which the optimal policy is acquired using reinforcement learning \cite{deudon2018learning}. Enhanced by deep learning techniques, the policies can be further parameterized through neural networks for learning better representation \cite{kool2018attention} \cite{peng2019deep}. 
The existing works have studied problems like the travelling salesman problem (TSP) and the vehicle routing problem where the searching space is finite, and therefore, they focus primarily on policy parameterization and training \cite{kool2018attention} \cite{peng2019deep}. The key challenge in solving the EST problem is that the searching space is unstructured: the Steiner points could be any subset of the Euclidean space. Therefore, as opposed to the TSP problem where the searching space is composed of all permutations of the nodes, the solutions to the EST problem cannot be readily enumerated. A straightforward method is to discretize the Euclidean space using grid networks by which the Steiner points are assumed to be on the joints. However, such a simple method is intuitively not optimal because it is unable to take account of any combinatorial properties of the Steiner trees. In this paper, we make an attempt to address the above challenges by presenting a reinforcement learning framework called Deep-Steiner for solving the EST problem. Our work in this paper can be summarized as follows:

\begin{itemize}
    \item \textbf{Space discretization.} Leveraging the unique properties of Steiner trees, we design methods for creating a compact candidate set for the Steiner points. The resulted candidate set is dynamically constructed during the incremental searching, thereby being superior to methods adopting fixed searching spaces (e.g., grid network). In addition, the complexity of our candidate set can be easily tuned through hyper-parameters, admitting a controllable efficacy-efficiency trade-off. 
    
    \item \textbf{Policy parameterization and training schemes.} We present methods for parameterizing the searching policy using attention techniques to dynamically update the point and graph embeddings. Our methods iteratively update the point and graph embeddings and then select a Steiner point until meeting the stopping criterion, which is different from previous methods that only compute the embeddings once. In addition, we design new reinforcement schemes for parameter training, where three stopping criteria are proposed for different training purposes.

    \item \textbf{Implementation and experiments.} Extensive experiments have been conducted to examine the proposed methods on different datasets and different training schemes. We have acquired encouraging results showing that the proposed method is non-trivially better than the baselines, which might be the first piece of evidence that demonstrates the potential of data-driven approaches for addressing combinatorial optimization problems with complex searching spaces. Our source code is made publicly available\footnote{\url{https://github.com/deepsteiner/Deepsteiner}}.
\end{itemize}

\section{Related work}
\textbf{Steiner tree in wireless networks.}
In the field of wireless network, many research problems are closely related to the Steiner tree problem. For example, the relay node placement problem can be reduced to the EST problem with bounded edge length \cite{lin1999steiner,cheng2008relay}. Furthermore, solving the EST problem is also a key to many other problems, including the problem of restoring the network connectivity \cite{senel2011relay}, the minimum connected dominating set problem \cite{min2006improving}, the problem of broadcast routing \cite{li2004energy}, and the minimum length multicast tree problem \cite{gong2015distributed}.

\textbf{Traditional methods.} Traditional approaches for solving the EST problem can be classified into three branches: exact algorithms, approximation algorithms, and heuristic methods. Warme \textit{et al.} propose the GeoSteiner algorithm \cite{warme1998spanning}, which is the most successful exact algorithm for the EST problem. Their algorithm is based on the generation and concentration of the full Steiner tree.
For heuristic algorithms, Thompson \textit{et al.} provide an edge insertion algorithm using local search methods \cite{Thompson1973TheMO}, and an improved version is designed later by Dreyer \textit{et al.} \cite{dreyer1998two}. 
Smith \textit{et al.} propose a heuristic algorithm running in $O(n\log n)$ based on the generation and concentration of the full Steiner tree \cite{smith1981n}. 
Bereta et al. propose a memetic algorithm to find optimal Steiner points \cite{bereta2019baldwin}. These heuristic algorithms can generate an approximate solution in a polynomial time, suggesting that good handcrafted rules are helpful to solve the EST problems. 

\textbf{Learning to solve combinatorial optimization problem.}
Using machine learning to learn heuristic algorithms opens new horizons for solving combinatorial optimization problems. Vinyals \textit{et al.} \cite{vinyals2015pointer} propose a supervised learning method and design the pointer network model to learn a heuristic for computing the Delaunay triangulation and solving TSP. Later, Bello \textit{et al.} \cite{Bello2017NeuralCO} propose a reinforcement learning method to solve TSP; following this idea, Kool \textit{et al.} \cite{kool2018attention} propose a deep reinforcement learning method based on the Transformer architecture. Motivated by the success on solving TSP, various combinatorial problems have been revisited by the machine learning community, e.g., the 3D bin packing problem \cite{hu2017solving}, the minimum vertex cover problem \cite{khalil2017learning}, and the maximum cut problem \cite{barrett2020exploratory,khalil2017learning}. More related works can be found in recent surveys \cite{mazyavkina2021reinforcement,bengio2021machine}.

\textbf{Steiner tree over graph vs Euclidean Steiner tree.} Being closely related to the EST problem, the Steiner tree over graph problem seeks to build the min-cost network using Steiner points from a given point set, implying the searching space is finite. Therefore, most of the existing techniques can be easily applied to the Steiner tree over graph problem. For example, Du \textit{et al.} \cite{du2021vulcan} propose a reinforcement learning method and Ahmed \textit{et al.} \cite{ahmed2021computing} study the same problem through supervised learning methods. Another related one is the rectilinear Steiner minimum tree problem, where the network is required to be composed of horizontal and vertical lines, and therefore, the grid network is a natural choice for creating the searching space \cite{chen2022reinforcement}. Unfortunately, such a simple method does not work well for the EST problem, as evidenced later in experiments.

\section{Problem setting}
Formally, the EST problem is defined as follows.
\begin{definition} [The Euclidean Steiner tree (EST) problem] 
Given a point set $S$ (called terminal set) in a two-dimensional Euclidean space $\mathbb{R}^2$, construct a tree with the minimum Euclidean distance to connect $S$, where points not in $S$ (called Steiner points) are allowed to use to construct the tree.
\end{definition}

Concerning a distribution $\D$ over the input instance $S$, we wish to learn a function $T$ that can compute a Steiner tree $T(S)$ over $S$ with the minimum length. As a statistical learning problem, we seek to minimize the true error:
\small
\begin{equation*}
    L(T) = \mathbb{E}_{S \sim \D}\Big[\len\big(T(S)\big)\Big]
\end{equation*}\normalsize
where $\len(T(S))$ denotes the tree length under the Euclidean distance.

\section{Deep-Steiner}

In this section, we present a reinforcement learning method called Deep-Steiner to learn the desired function $T$.

\begin{algorithm}[t]
\caption{Deep-Steiner} \label{A1}
\label{alg:algorithm}
\textbf{Input}: $S$ and $p_{\theta}$\\
\textbf{Output}: A Steiner tree associated with $S$
\begin{algorithmic}[1]
\STATE $I_0 = S, i=1$ 
\WHILE{$\len(\MST(I_i))\leq \len(\MST(I_{i-1})) $ and $i \leq |S| -2$}
\STATE Generate candidate set $C_i$ based on $I_{i-1}$
\STATE $v^* = \argmax_{v \in C_i} p_{\theta}(v|I_{i-1})$, $I_i=I_{i-1} \cup \{v^*\}$, $i=i+1$
\ENDWHILE
\STATE \textbf{return} $\MST(I_{i-1})$
\end{algorithmic}
\end{algorithm}

\subsection{Overall structure}
Notably, supposing that the optimal Steiner points have been identified, the optimal solution must be a minimum spanning tree over the Steiner points plus the terminal nodes. Therefore, the key part is to determine the Steiner points. The overall framework of Deep-Steiner is conceptually simple, as shown in Algorithm \ref{A1}. Given the input instance $S$, Deep-Steiner decides the Steiner points in an iterative manner. In each iteration $i$, based on the current selected points $I_{i-1} \subseteq \mathbb{R}^2$ including both the terminal points and Steiner points, a new Steiner point is selected from a certain candidate set $C_i \subseteq \mathbb{R}^2$ by a policy $\pi: C_i \rightarrow \mathbb{R}^2$, with $I_0=S$ being the initial input. In particular, the policy selects the best node from the candidate set $C_i$ according to a distribution $p_\theta(v|I_{i-1})$ over $v \in C_i$ (conditioned on $I_{i-1}$), where $p_\theta$ is parameterized by deep neural networks. In completing the framework, we first present methods for generating the candidate set (Section \ref{subsec: candidate}) and then present the design of $p_\theta$ (Section \ref{subsec: policy}). Finally, we discuss training methods for computing the parameters $\theta$ involved in $p_\theta$ (Section \ref{subsec: training}).

\subsection{Candidate set $C_i$}
\label{subsec: candidate}
We now present methods for constructing the candidate set $C_i$ based on the current selected points $I_{i-1}$. To eschew searching the continuous space $\mathbb{R}^2$, one straightforward idea is to use the grid network, which has been widely adopted in existing methods \cite{saeed2019path,ammar2016relaxed,chen2022reinforcement}. Nevertheless, the grid pattern is not informed by the structure of the Steiner trees and therefore can lead to suboptimal performance, as shown in the experiments later. For a better candidate space, we are inspired by the following property of the Steiner trees. 
\begin{lemma} \cite{gilbert1968steiner} {\label{degree}}
The Steiner points in the Steiner minimal tree must have a degree of three, and the three edges incident to such a point must form three 120 degree angles.
\end{lemma}
The above property suggests that the optimal Steiner points must be on the 120 degree-arc between two points in $\mathbb{R}^{2}$, which has greatly reduced the searching space. We denote such arcs as Steiner arcs.
\begin{definition} [Steiner arc]
Given two points $a \in \mathbb{R}^2$ and $b \in \mathbb{R}^2$, the set $\arc_{a, b} = \{x: x\in \mathbb{R}^2, \angle axb = 120\}$ is the Steiner Arc induced by $a$ and $b$.
\end{definition}
With the above concept, the initial space we acquire consists of all the Steiner arcs between the points in $I_{i-1}$. As such a space is still continuous, we divide the arc into $k^{*}$ equal-size parts, for some $k^* \in \mathbb{Z}$, and select the endpoints to acquire an enumerable space. Formally, the resulted candidates associated with one arc $\arc_{a, b}$ are given as follows. 
\small
\begin{align*}
\label{eq: k-partition}
&\arc_{a, b}^{k^{*}} = \\
&\Big\{x_1,...,x_{k^{*}} : x_i \in \arc_{a, b}, |x_{i-1}-x_{i}|=|x_{i}-x_{i+1}|, x_0 = a, x_{k^{*}+1}=b, i \in [k^{*}] \Big\}
\end{align*}\normalsize

\begin{figure}[t!]
{\includegraphics[width=\textwidth]{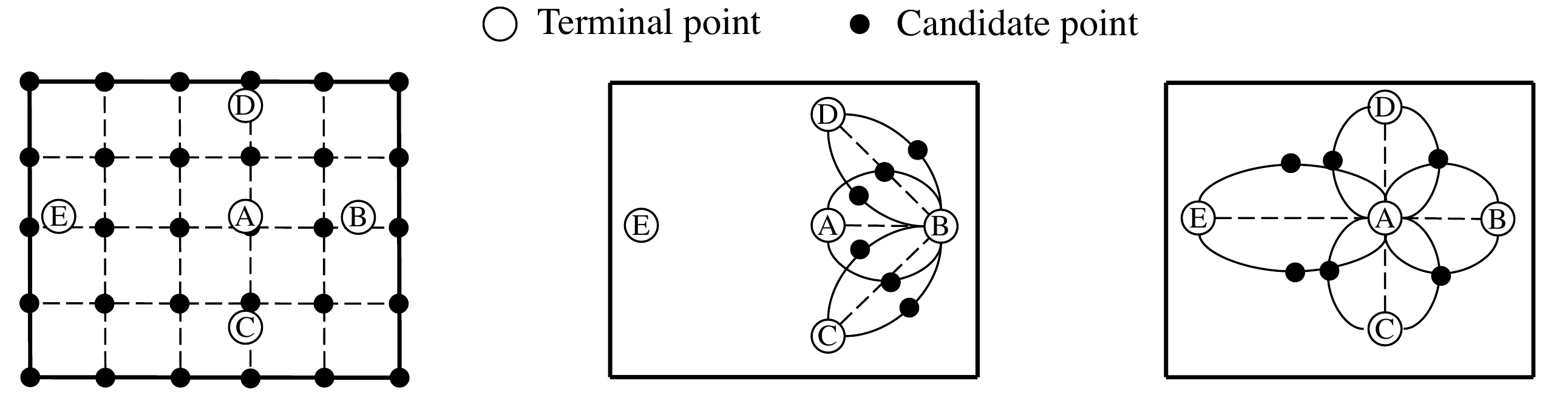}}
\caption{The left figure illustrates the candidate points based on the grid network. The middle figure shows the candidate points generated based on the Steiner arcs between node $B$ and its nearest neighbors; the right figure presents the candidate points generated by the MST-based methods.} \label{fig: candidate}
\end{figure}

Enumerating over all possible pairs in $I_{i-1}$, the total candidate points are in the order of $O(k^{*}\cdot |I_{i-1}|^2)$, which is still infeasible when being involved in training deep architectures (Section \ref{subsec: policy}). Therefore, we adopt the following two methods to further reduce the searching space: 

\begin{itemize}
    \item \textbf{KNN-based candidate space.} Notice that the Steiner minimal tree has the optimality condition that each subgraph tree is also a minimum spanning tree over the involved points. Intuitively, for two points in $I_{i-1}$ that are not local neighbors, their Steiner arc cannot include any Steiner points in the optimal solution to the input instance $S$.  Therefore, for each point $v$ in $I_{i-1}$, only the Steiner arcs between $v$ and its nearest neighbors are selected. Let $N_v^{k^{'}} \subseteq I_{i-1}$ be the set of the $k^{'} \in \mathbb{Z}$ nearest neighbors of $v$ in  $I_{i-1}$. We have the following candidate space:
    \begin{equation}
        C_i = \bigcup_{v \in I_{i-1}} \bigcup_{u \in N_v^{k^{'}}} \arc_{v, u}^{k^{*}}
    \end{equation}
    
    \item \textbf{MST-based candidate space.} Following the same intuition of the KNN-based method, we now seek to eliminate unqualified Steiner arcs by using the minimum spanning tree $\MST(I_i)$ over $I_i$. That is, for each pair of the nodes in $I_{i-1}$, we only select the Steiner arc $\arc_{u,v}$ when edge $(u,v)$ appears in $\MST(I_i)$. Formally, we have the following space:
    \begin{equation}
        C_i = \bigcup_{(u,v) \in \MST(I_{i-1})} \arc_{v, u}^{k^{*}}
    \end{equation}
\end{itemize}
One can easily see that the candidate space resulted from the above methods is linear in $|I_{i-1}|$ and determined by some hyperparameters $k^*$ and $k^{'}$ that control the complexity-efficient trade-off. Such a trade-off will be experimentally studied later. An illustration of the candidate points generated by different methods is given in Figure \ref{fig: candidate}.

\subsection{Policy $\pi$ and distribution $p_{\theta}$}
\label{subsec: policy}

Given the candidate set $C_i$ in each iteration of Algorithm \ref{A1}, our policy computes one Steiner point based on a parameterized distribution $p_{\theta}(v|I_{i-1})$, with the hope that a higher value of $p_{\theta}(v|I_{i-1})$ translates the better quality of the final Steiner tree. To this end, the parameters $\theta$ will be trained using reinforcement learning (Section \ref{subsec: training}). In what follows, we present our design of $p_{\theta}(v|I_{i-1})$. The overall design of $p_{\theta}$ has two modules: the encode computes the hidden embedding of each point in $C_i$ as well as the entire graph; the decoder translates the embeddings into a distribution over $C_i$.

\textbf{Encoder.} Given the candidate $C_i$ and the currently selected points $I_{i-1}$, where each point is represented by its coordinates, we first compute the hidden representations following the framework proposed in \cite{kool2018attention} to capture useful latent relationships between the points in terms of building a Steiner minimal tree. 

Let $X=C_i \cup I_{i-1} =\{x_1,...,x_n\}$ be the entire point set. For each $x_j \in X$, the embedding is computed through the attention mechanism \cite{vaswani2017attention} composed of a sequence of $L \in \mathbb{Z}$ neural layers, where its $d$-dimensional hidden representation at layer $l \in [L]$ is denoted by $\vec{A}_{j,l} \in \mathbb{R}^{d \times 1}$. Initially, we have $\vec{A}_{j,0} =\vec{W}^{(0)} \cdot x_j + \vec{b}^{(0)}$ with $\vec{W}^{(0)}\in \mathbb{R}^{d \times 2}$,  and $\vec{b}^{(0)} \in \mathbb{R}^{d \times 1}$ being learnable parameters. To obtain $\vec{A}_{j,l+1}$ from $\vec{A}_{j,l}$, we first use multi-head attention \cite{vaswani2017attention} to aggregate the information from the neighbors of $x_j$:
\small
\begin{equation*}
\MHA_{j, l+1}(\vec{A}_{1, l},\vec{A}_{2, l},...,\vec{A}_{n, l}) = \sum_{m=1}^M \vec{W}^{(1)}_{m,l} \cdot  \vec{A}_{j, m}^{'}
\end{equation*}\normalsize
which is composed of $M \in \mathbb{Z}$ self-attentions $\vec{A}^{'}_{j,m}$ \cite{vaswani2017attention} computed by
\small
\begin{equation*}
     \vec{A}_{j, m}^{'} = \sum_{k=1}^{n} \frac{  \exp(\mathbf{q}_{j, m}\tran \mathbf{k}_{k, m}/\sqrt{d_s}) \cdot \mathbf{v}_{k, m} }{\sum_{k^{'}=1}^{n}\exp(\mathbf{q}_{j, m}\tran \mathbf{k}_{k^{'}, m}/\sqrt{d_s})}
\end{equation*}\normalsize
where $d_s=d/M$ is the normalizer, $\mathbf{q}_{j, m} \in \mathbb{R}^{d_s \times 1}$, $\mathbf{k}_{j, m} \in \mathbb{R}^{d_s \times 1}$, and $\mathbf{v}_{j, m} \in \mathbb{R}^{d_s \times 1}$ are hidden vectors obtained through linear transformation from $\vec{A}_{j, l}$:
\small
\begin{align*}
    \mathbf{q}_{j, m} = \vec{W}^{(2)}_{m, l+1} \cdot \vec{A}_{j, l}, ~~\mathbf{k}_{j, m} = \vec{W}^{(3)}_{m, l+1} \cdot  \vec{A}_{j, l}, ~~\mathbf{v}_{j, m} = \vec{W}^{(4)}_{m, l+1} \cdot  \vec{A}_{j, l}
\end{align*}\normalsize
with $\vec{W}^{(1)}_{m, l+1} \in \mathbb{R}^{d \times d_s}$,  $\vec{W}^{(2)}_{m, l+1} \in \mathbb{R}^{d_s \times d}$, $\vec{W}^{(3)}_{m, l+1} \in \mathbb{R}^{d_s \times d}$, and $\vec{W}^{(4)}_{m, l+1} \in \mathbb{R}^{d_s \times d}$ being learnable parameters.
With the transmission of $\MHA$, the representation $\vec{A}_{i, l+1}$ of point $x_j$ at the next layer $l+1$ is finally obtained through batch normalization and one feed-forward layer:
\small
\begin{align*}
\hat{\vec{A}}_{j,l} &= \BN_{l+1} \big(\vec{A}_{j, l}+ \MHA_{j, l+1}(\vec{A}_{1, l},\vec{A}_{2, l},...,\vec{A}_{n, l})\big)\\
\vec{A}_{j,l+1} & = \BN_{l+1} \big(\hat{\vec{A}}_{j, l}+ \FF_{l+1}(\hat{\vec{A}}_{j, l})\big)
\end{align*}\normalsize
where $\BN$ is the standard batch normalization process \cite{ioffe2015batch} and $\FF$ is the standard feed-forward network with ReLu activations \cite{Goodfellow-et-al-2016} of dimension $512$. Finally, the graph embedding $\vec{A}_G \in \mathbb{R}^{d}$ is calculated as the average sum of the final embeddings over $I_{i-1}$ rather than over $X$, which is different from \cite{kool2018attention}:
\small
\begin{equation*}
   \vec{A}_G = \frac{1}{|I_{i-1}|} \sum_{j: x_j \in I_{i-1}} \vec{A}_{j, L}
\end{equation*}\normalsize

\textbf{Decoder.} With the final point embeddings $\vec{A}_{j,L}$ and the graph embedding $\vec{A}_G$, the encoder derives a parameterized distribution over $C_i$ by combining the embeddings through exponential families:
\begin{equation*}
    p_\theta( x_j \in C_i | I_{i-1}) = \frac{\exp(\vec{q}\tran\cdot  \vec{k}_{j}/\sqrt{d_s})} {\sum_{k: x_k \in X_i\cap C_i}\exp(\vec{q}\tran\cdot  \vec{k}_{k}/\sqrt{d_s})}
\end{equation*}
where the hidden representation $\vec{q} \in \mathbb{R}^{d_s \times 1} $ and $\vec{k}_j \in \mathbb{R}^{d_s \times 1}$ are obtained via linear transformation based on the embeddings:
\begin{align*}
   \vec{q} = \vec{W}^{(5)} \cdot \vec{A}_G ~~\text{and}~~   \vec{k}_j = \vec{W}^{(6)} \cdot \vec{A}_{j, L} 
\end{align*}
where $\vec{W}^{(5)}  \in \mathbb{R}^{d_s \times d} $ and $\vec{W}^{(6)}  \in \mathbb{R}^{d_s \times d} $ are learnable.

\textbf{Parameter space.} In summary, the proposed model has three hyperparameters: $M \in \mathbb{Z}$ controlling the complexity of multi-head attention, $L \in \mathbb{Z}$ determining the number of layers in the encoder, and $d \in \mathbb{Z}$ specifying the hidden dimensions. On top of that, the distribution $p_{\theta}$ is parameterized by a collection of learnable matrices, including $\vec{W}^{(0)}$, $\vec{b}^{(0)}$, $\vec{W}^{(i)}_{m,l}$ ($i \in [1, 4]$, $m \in [M]$ and $l \in [L]$), $\vec{W}^{(5)}$, and $\vec{W}^{(6)}$. 

\subsection{Training} 
\label{subsec: training}

\begin{algorithm}[t]
\caption{REINFORCE with Rollout Baseline} \label{A2}
\textbf{Input}:  batch size $B$, significance $\alpha$ \\
\textbf{Output}: parameter $\theta_{bs}$
\begin{algorithmic}[1]
\STATE initial parameter $\theta$, $\theta_{bs} \leftarrow \theta$
\FOR{each epoch}
\FOR{each batch training}
\STATE Sample a batch of instance $\{S_1,...,S_B\}$
\FOR{$j \in [B]$} 
\STATE $I_0 = S_j$, $i=1$ 
\WHILE{stopping criterion is not met}
\STATE Generate candidate set $C_i$ based on $I_{i-1}$
\STATE $v^{'} \sim p_{\theta}(v|I_{i-1})$, $I_i=I_{i-1} \cup \{v^{'}\}$, $i=i+1$
\ENDWHILE
\STATE $T_{\theta}(S_i)= \MST(I_{i-1})$ 
\STATE $I_0 = S_j$, $i=1$ 
\WHILE{stopping criterion is not met}
\STATE Generate candidate set $C_i$ based on $I_{i-1}$
\STATE $v^* = \argmax_{v \in C_i} p_{\theta_{bs}}(v|I_{i-1})$, $I_i=I_{i-1} \cup \{v^*\}$, $i=i+1$
\ENDWHILE
\STATE $T_{\theta_{bs}}(S_i)= \MST(I_{i-1})$ 
\ENDFOR
\STATE $\bigtriangledown \mathcal{L} \leftarrow \sum_{i=1}^B (\len(T_{\theta}(S_i))- \len(T_{\theta_{bs}}(S_i)))\cdot \bigtriangledown_\theta \log p_{\theta}(T_{\theta}(S_i)| S_i)$ 
\STATE $\theta \leftarrow$ Adam$(\theta,\bigtriangledown \mathcal{L})$
\ENDFOR
\IF{t-test($\theta, {\theta_{bs}}$, $\alpha$) passes}
\STATE ${\theta_{bs}}\leftarrow \theta$ 
\ENDIF
\ENDFOR
\end{algorithmic}
\end{algorithm}

We now present methods for computing the parameter using REINFORCE with baseline \cite{williams1992simple,kool2018attention}. The framework is given in Algorithm \ref{A2}. Each epoch consists of a sequence of batch training. In each batch training (line 4-20), given a batch size $B \in \mathbb{Z}$, we sample a collection of instances, and for each instance, we predict the Steiner trees based respectively on the current parameter $\theta$ and the best parameter $\theta_{bs}$. With the predicted Steiner trees, we obtain a new parameter $\theta_{new}$ by maximizing
\begin{align}
 \sum_{i=1}^B (\len(T_{\theta}(S_i))- \len(T_{\theta_{bs}}(S_i)))\cdot \bigtriangledown_\theta \log p_{\theta}(T_{\theta}(S_i)| S_i)
\end{align}
using the Adam algorithm \cite{kingma2015adam}, where $p_{\theta}(T_{\theta}(S_i)| S_i) = \prod_{j} p_{\theta}(I_{j}| I_{j-1})$ is likelihood of the prediction and $I_j$ is the sequence of points used to construct $T_{\theta}(S_i)$. The above optimization implies that the model should produce a Steiner tree with small length and large likelihood. At the end of each epoch, the parameter will be updated only if it passes the paired t-test with $\alpha=0.05$ on $10,000$ validation instances to examine whether or not the performance improvement is statistically significant \cite{hsu2014paired}. Now the only part left is the process to generate predictions during the training process (line 6-17 in Algorithm \ref{A2}), where we keep selecting Steiner points but use different stopping criteria for different training purposes. We consider three stopping criteria:
\begin{itemize}
    \item \textbf{First-increment.} Similar to the inference process (Algorithm \ref{A1}), we stop adding Steiner points if the resulted minimum spanning tree becomes larger instead of smaller. The rationale behind such a strategy is that parameter updating would become useless if the current tree is nowhere close to the optimal one. 
    \item \textbf{First-selection.} Under this stopping criterion, we stop adding Steiner points after selecting one Steiner point, which means that we focus on training the policy to select the nodes that are locally optimal. Since our inference process runs in a greedy manner, local optimality is necessary to achieve nontrivial performance compared to random methods. 
    \item \textbf{All-selection.} Under such a method, we always select $|S|-2$ Steiner points, which is motivated by the well-known fact that the Steiner minimal tree over $|S|$ points can have at most $|S|-2$ Steiner points \cite{gilbert1968steiner}.
\end{itemize}

\section{Experiment}
In this section, we present our empirical studies.
\subsection{Experiment setting}

\textbf{Model setting.} We examine our method with different candidate generation methods (Section \ref{subsec: candidate}) and different stopping criteria (Section \ref{subsec: training}). Following the standard practice \cite{qiu2020pre}, we initialize the parameters using a pre-trained TSP-20 model  \cite{kool2018attention}. Balancing between quality and computational cost, the best hyperparameters are set as follows. $k^* = 9$ for arc partition; $k^{'} = 3$ for KNN-based candidate space; $d = 128, L = 5, M = 8$ for the attention model; the batch size $B$ is $32$.

\textbf{Datasets.} For synthetic dataset $(n,\D)$, each instance is generated by randomly selecting $n$ points following a certain distribution $\D$ over $\mathbb{R}^2$. Our experiments involve three synthetic datasets: \small
\[D_1=(10, U(0,1)^2),D_2=(20, U(0,1)^2),D_3=(10, N(0.5,0.2)^2),\]\normalsize
where $U$ is the uniform distribution and $N$ is the Gaussian distribution. Such samples have been widely adopted in existing works \cite{kool2018attention,joshi2019efficient}. In addition, we adopt the DEG-10 dataset \cite{wong2008scalable}, which is a benchmark used in the 11th DIMACS Implementation Challenge \cite{11ch}.

\textbf{Baseline methods.} The optimal solution is calculated by Geo-Steiner \cite{warme2003geosteiner}. The second baseline outputs the minimum spanning tree of the input instance, which is a simple and effective approximation to the Steiner minimal tree. In addition, we include a highly effective heuristic method proposed by Bereta \textit{et al.} \cite{bereta2019baldwin}, which is an iterative framework and can often produce near-optimal solutions provided with a sufficient number of iterations. Finally, two random methods are adopted as baselines. Rand-1 samples $r \in \mathbb{Z}$ random Steiner points from the distribution $\D$, where $r$ is sampled from $[0, |S|-2]$. Rand-2 is the same as Algorithm 1 except that the Steiner points are randomly selected from the candidate set; such a method is used to prove that our policy is indeed nontrivial. 

\textbf{Training and testing.} Our experiments were executed on Google Colab with NVIDIA Tesla V100. The training size is $10,240$. For each method, we examine its performance by comparing the predicted Steiner trees to the optimal solution over $10,000$ testing instances, i.e., $\frac{\len(T_{\theta}(S))-\len(T_{\opt}(S))}{\len(T_{\opt}(S))}$.

\begin{table}[t!] 
\centering
\renewcommand{\arraystretch}{1.0} 
\caption{\textbf{Main results.} Each cell shows the performance of one method on one dataset, together with the standard deviation. The missing data means that the training time of one epoch is more than 2.5 hours, which is considered to be impractical.}\label{EvaT}
%\small
\begin{tabular}{@{}l @{\hspace{6mm}} r @{\hspace{4mm}} r @{\hspace{4mm}} r @{}}
        %\toprule
        Methods & $D_1$ & $D_2$ & $D_3$\\
        \midrule
        Geo-Steiner  & 0  & 0 & 0\\
        Bereta's method  (10 iterations)  &1.67 $\pm$0.02\% &2.43 $\pm$0.01\% &1.67 $\pm$0.02\%\\
        Bereta's method  (20 iterations)  &1.08 $\pm$0.01\% &1.92 $\pm$0.01\% &1.13 $\pm$0.02\%\\
        Bereta's method  (50 iterations)   &0.59 $\pm$0.01\%& 1.18 $\pm$0.01\% &0.61 $\pm$0.01\%\\
        Minimum spanning tree  & 3.15 $\pm$0.03\% &3.19 $\pm$0.01\% &3.06 $\pm$0.02\%\\
        Rand-1  &25.05 $\pm$3.31\% &24.08 $\pm$2.06\% & 48.44 $\pm$19.1\%\\
        \midrule
        KNN+First-selection  & 2.41 $\pm$0.02\%  & 2.13 $\pm$0.01\%  &2.62 $\pm$0.02\%\\
        KNN+First-increment  & 2.94 $\pm$0.02\%  &  & 2.82 $\pm$0.02\%\\
        KNN+All-selection  & 3.02 $\pm$0.02\% &  & 2.92 $\pm$0.02\%\\
        KNN+Rand-2   &3.64 $\pm$0.03\% & 3.24 $\pm$0.01\% &3.45 $\pm$0.03\%\\
        \midrule
        MST+First-selection & 1.39 $\pm$0.01\%  & 2.46 $\pm$0.01\%  & 1.77 $\pm$0.01\%\\
        MST+First-increment   & 2.30 $\pm$0.02\%  & 3.01 $\pm$0.01\%  &  2.07 $\pm$0.01\%\\
        MST+All-selection  & 2.95 $\pm$0.02\%  &  & 2.92 $\pm$0.02\%\\
        MST+Rand-2  &4.20 $\pm$0.04\% & 3.62 $\pm$0.02\%  & 3.89 $\pm$0.04\%\\
        \midrule
        Grid+First-selection  &2.98 $\pm$0.02\%  &3.16 $\pm$0.01\%  & 2.88 $\pm$0.02\%\\
        Grid+First-increment&3.08 $\pm$0.02\%  &3.16 $\pm$0.01\%  & 2.93 $\pm$0.02\%\\
        Grid+All-selection &3.07 $\pm$0.02\% & & 2.91 $\pm$0.02\%\\
        Grid+Rand-2 &3.31 $\pm$0.03\% &3.21 $\pm$0.01\% &3.23 $\pm$0.03\%\\
        %\bottomrule
\end{tabular}%\normalsize
\vspace{-6mm}
\end{table}

\subsection{Result and analysis}
\textbf{Overall observations.} Tables 1 and 2 show the performance in terms of effectiveness and efficiency. The results confirm that our method is non-trivially better than Rand-1 and minimum spanning trees, and furthermore, it is comparable to the best heuristic and can generate solutions close to the optimal ones. Compared to Berata's method, the main advantage of our method lies in time efficiency (as shown in Table 2), which suggests that methods based on reinforcement learning are promising in dealing with NP-hard combinatorial optimization problems with complex searching spaces.

\begin{table}[t!] 
\renewcommand{\arraystretch}{1.1} 
\centering
\caption{\textbf{Running time.} Each cell shows the average time cost to generate one prediction.} \label{RT}
    \begin{tabular}{@{}l @{\hspace{5mm}} r @{\hspace{5mm}} r @{\hspace{5mm}} r @{}}
        %\toprule
        Methods & $D_1$& $D_2$& $D_3$\\
        %& $\D=U(0,1)^2$ & $\D= U(0,1)^2$ & $\D=N(0.5,0.2)^2$\\
        \midrule
        Geo-Steiner &  211.2ms& 303.2ms&  205.2ms\\
        Bereta's method (10 iterations) & 118.8ms& 331.7ms&  125.2ms\\
        Bereta's method (20 iterations)& 310.0ms & 898.5ms&  315.4ms\\
        Bereta's method (50 iterations) & 1021.5ms& 2217.1ms&  945.6ms\\
        Minimum spanning tree  & 1.6ms& 5.8ms&  1.7ms\\
        Rand-1 & 2.8ms& 10.7ms&  2.9ms\\
        \midrule
        KNN+First-selection &  200.6ms& 285.1ms&  148.3ms \\
        KNN+Rand-2 &  9.4ms&  41.9ms &  8.7ms\\
        \midrule
        MST+First-selection  &  142.1ms&  254.9ms &  130.7ms\\
        MST+Rand-2 &  14.3ms&  22.3ms &  12.9ms\\
        \midrule
        Grid+First-selection  & 39.3ms& 72.1ms&  35.8ms \\
        Grid+Rand-2  & 7.0ms& 20.4ms&  7.0ms \\
        %\bottomrule
    \end{tabular}
    \vspace{-6mm}
\end{table}

\textbf{On candidate generation methods.} According to Table 1, the proposed methods are clearly better than grid based methods in generating good candidate points. In addition, MST-based methods are better than the KNN-based on small graphs but not on large graphs. The main reason is that the candidate set generated by KNN-based methods is larger than that of the MST-based methods, which means that, compared to the MST-based methods, KNN-based methods need more training epochs to get converged (especially on larger graphs) but can have a better performance once it is converged, for example, on small graphs. Indeed, we observed that both methods have almost converged on small graphs, while the loss of the KNN-based methods was still decreasing after 100 training epochs on large graphs.

\textbf{On stopping criteria.}
According to Table 1, our proposed stopping criteria perform non-trivially better than the Rand-2, which confirms that our models indeed learn to improve the policy during training. An interesting observation is that First-selection has the best performance among the possible stopping criteria. One plausible reason is that in generating predictions in the training phase, compared to First-selection, First-increment and All-selection tend to construct larger trees on which the loss function is defined, which means that the corresponding models seek to learn a policy that can draw a Steiner tree, as opposed to the case of the First-selection where the model simply wants to learn how to select the best local Steiner point. For All-selection, it is less optimal also because it often forces the model to select some unnecessary candidates, as most of the Steiner minimal trees do not have $|S|-2$ Steiner points. 

\textbf{Ability to generalize between distributions.}  In order to investigate the generalization performance between distributions, we train the MST+First-selection on one dataset and evaluate its performance on different datasets. The result of this part is given in Table 3. The main observation is that even with distributions shifts, our method can still generate Steiner trees that are non-trivially better than the minimum spanning trees, which suggests that the embeddings we acquire do not heavily rely on the points distributions.

\begin{table}[t!] 
\renewcommand{\arraystretch}{1.2} 
\centering
    \caption{\textbf{Generalization performance.} Each cell shows the performance of one method generalized to one dataset, together with the standard deviation)} \label{GT}
    \begin{tabular}{@{}l @{\hspace{4mm}} r @{\hspace{4mm}} r @{\hspace{4mm}} r @{\hspace{4mm}} r @{}}
        %\toprule
        Methods & $D_1$& $D_2$& $D_3$& DEG\\
        \midrule
        Model trained on $D_1$ &1.39 $\pm$0.01\% & 2.72 $\pm$0.01\% & 1.63 $\pm$0.01\% & 1.47 $\pm$0.01\%\\
        Model trained on $D_2$ &2.11 $\pm$0.02\% & 2.46 $\pm$0.01\% & 2.43 $\pm$0.05\% & 2.14 $\pm$0.02\%\\
        Model trained on $D_3$ &1.75 $\pm$0.01\% & 2.81 $\pm$0.01\% & 1.77 $\pm$0.01\% & 1.81 $\pm$0.02\%\\
        Minimum spanning tree & 3.15 $\pm$0.03\%  & 3.19 $\pm$0.01\% & 3.06 $\pm$0.02\% & 3.29 $\pm$0.02\%\\
        %\bottomrule
    \end{tabular}
    \vspace{-0mm}
\end{table}

\section{Conclusion} 
In this paper, we propose Deep-Steiner, a deep reinforcement learning method, to solve the EST problem. In particular, we design space discretization methods based on the unique properties of minimal Steiner trees, policy parameterization methods using attention techniques, and new training schemes with different stopping criteria. As evidenced by experiments, our method can generate decent solutions without incurring high computation and memory costs compared to traditional algorithms. One future work is to explore new representation methods in order to improve the performance on small graphs. In addition, it remains unknown how to overcome the memory issues for handling large instances, which is another interesting future work.
\bibliographystyle{splncs04}
\bibliography{samplepaper.bib}
\end{document}